\documentclass[10pt, conference, compsocconf]{IEEEtran}
\usepackage{cite}

\ifCLASSINFOpdf
\else
\fi
\usepackage[cmex10]{amsmath}
\usepackage{algorithmic}
\usepackage[caption=false,font=footnotesize]{subfig}
\usepackage{stfloats}

\usepackage{url}

\usepackage[bookmarks=false]{hyperref}
\usepackage{graphicx}
\usepackage{tikz}
\def\checkmark{\tikz\fill[scale=0.4](0,.35) -- (.25,0) -- (1,.7) -- (.25,.15) -- cycle;} 

\newcommand{\etal}{\textit{et al}.}

\begin{document}

\title{SpotNet: Self-Attention Multi-Task Network for Object Detection}


\author{\IEEEauthorblockN{Hughes Perreault, Guillaume-Alexandre Bilodeau, Nicolas Saunier}
\IEEEauthorblockA{
Polytechnique Montréal\\
Montréal, Canada\\
Email: \{hughes.perreault, gabilodeau, nicolas.saunier\}@polymtl.ca}
\and
\IEEEauthorblockN{Maguelonne Héritier}
\IEEEauthorblockA{
Genetec\\
Montréal, Canada\\
Email: mheritier@genetec.com}
}

%


\maketitle

\begin{abstract}

Humans are very good at directing their visual attention toward relevant areas when they search for different types of objects. For instance, when we search for cars, we will look at the streets, not at the top of buildings. The motivation of this paper is to train a network to do the same via a multi-task learning approach. To train visual attention, we produce foreground/background segmentation labels in a semi-supervised way, using background subtraction or optical flow. Using these labels, we train an object detection model to produce foreground/background segmentation maps as well as bounding boxes while sharing most model parameters. We use those segmentation maps inside the network as a self-attention mechanism to weight the feature map used to produce the bounding boxes, decreasing the signal of non-relevant areas. We show that by using this method, we obtain a significant mAP improvement on two traffic surveillance datasets, with state-of-the-art results on both UA-DETRAC and UAVDT. 

\end{abstract}


\IEEEpeerreviewmaketitle

\section{Introduction}

There is increasing interest in automatic road user detection for intelligent transportation systems, advanced driver assistance systems, traffic surveillance, etc. Road user detection has its own set of challenges and difficulties, such as the high speed of some road users, the frequent occlusion between them and the small size of road users appearing afar. Despite huge improvements in the last years thanks to advancements in deep learning-based methods, results still need to be improved for reliable practical applications.

Recently, a new family of object detectors were proposed based on keypoint detection rather than based on bounding box classification~\cite{law2018cornernet,duan2019centernet,zhou2019objectsaspoints}. This approach presents several advantages, including not having to manually design anchor boxes and having to process fewer candidate boxes. Detecting objects in this way is deceptively simple and elegant, and quite fast. It yields state-of-the-art accuracy results on several datasets. Therefore, in this work, we build upon CenterNet~\cite{zhou2019objectsaspoints} by designing a novel convolutional neural network (CNN) model that directs its attention towards the areas of interest and thus decreases the probability of having false detections in incongruous areas. 

Our contributions are: 1)~a self-attention mechanism based on multi-task learning (object detection and segmentation) and 2)~a semi-supervised training method that capitalizes on automatic foreground/background segmentation annotations.  

\begin{figure}[t]
\begin{center}
\includegraphics[width=1\linewidth]{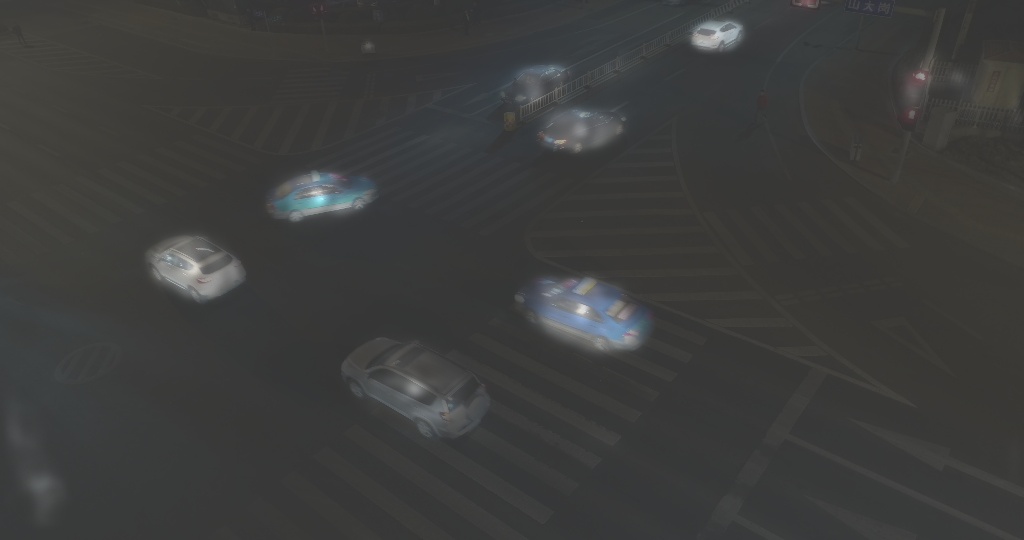}
\end{center}
   \caption{A visualisation of the attention map produced by SpotNet on top of its corresponding image, from the UAVDT~\cite{du2018unmanned} dataset.}
\label{illuminated}
\end{figure}

The idea of attention and self-attention has been around for some time now, most notably in image captioning~\cite{xu2015show} and natural language processing (NLP)~\cite{vaswani2017attention}. In those works, neural networks are trained to learn which parts of the input are the most important to solve the task. But they do so progressively, using recurrent neural networks. Can a simple CNN learn which areas it should use to increase its visual attention? In this work, we show that it is indeed possible and beneficial for object detection by using a semi-supervised training approach and multi-task learning. The network is trained for both object detection and foreground/background segmentation, the latter being also used to weight object detection feature maps. Indeed, the foreground/background segmentation is used in an internal attention mechanism that gives more weight to areas useful for detection. In figure~\ref{illuminated}, we can visualize what the network learns from this approach, that is to concentrate the keypoint search on areas where there are indeed road users, and therefore reducing the response of any other neuron. One can see this process as shining a spotlight on relevant areas and dimming the lights everywhere else. Therefore, we named our method, SpotNet.

This attention approach is particularly beneficial for keypoint-based methods since we are globally looking for keypoints on the whole image at the same time, and not just classifying the object in a cropped bounding box. However, a question remains. How can we train such a self-attention process? Typically, object detection datasets do not provide the segmentation ground-truth since it is very costly and time-consuming to produce. Instead, we rely on classical computer vision techniques to generate automatic pixel-wise annotation labels and on datasets providing video sequences instead of single frames to train the network. In the case of fixed camera video sequences, we successfully employ a background subtraction method to obtain the automatic annotations, while in the case of moving camera video sequences, we rely on dense optical flow for the same purpose. 

Although we use imperfect foreground/background segmentation annotations, we can train a network to produce quality segmentation maps by using multi-task learning. The detection and segmentation tasks are trained jointly by sharing all the parameters of the backbone network. Both tasks are mutually beneficial. Indeed, by producing a better segmentation, the object detection task benefits from a better attention mechanism. And by producing better object detection, the parameters of the backbone network get better at recognizing the features of interest from the images to improve the segmentation maps. We validated our method on two popular traffic scene datasets, and we show that our method is the state-of-art on these datasets by improving significantly the performance of the base network (CenterNet). 

\section{Related Work}

Object detection as meant in this paper is the task of drawing a rectangular bounding box around objects of interest in an image, as well as producing a class label for each box. All state-of-the-art object detection methods have been based on deep learning since its rise. They can broadly be split up into two main categories, two-stage and one-stage methods.

\textbf{Two-stage methods} divide the task of object detection into two steps, producing a set of object candidates, and then computing a score, a label and a coordinate offset for each box. The first deep learning-based method was R-CNN~\cite{rcnn_Girshick_2014_CVPR}, which used an external method to produce box candidates, namely selective search~\cite{uijlings2013selective}. It then passed each candidate in a CNN to compute features for each box. A classification is done on those features by a SVM afterwards. Fast R-CNN~\cite{Girshick_2015_ICCV_fast} aimed to increase the speed of R-CNN by passing the whole image through a CNN once, and afterwards just cropped the relevant parts of the feature map for each box candidate for classification. Faster R-CNN~\cite{ren2015faster} is a further improvement that introduced the RPN, a region proposal network that shares most of its parameters with the classification and regression parts, making it even faster and more efficient than its predecessors. RFCN~\cite{RFCN_NIPS2016_6465} further builds upon Faster R-CNN by learning to detect and classify parts of objects and then using a grid of parts to vote on each object. Cascade R-CNN~\cite{cai2018cascade} addresses the problems of the mismatch between the minimum IOU (Intersection over union) used to evaluate during inference, and the minimum IOU used to select a positive sample during training. They also address overfitting during training by training while progressively increasing the IOU thresholds. 

\textbf{One-stage methods} aim to reduce the processing time of two-stage methods by removing the candidate proposal phase and by detecting objects directly from the feature map. The first one-stage method was YOLO~\cite{redmon2016you_yolo} which divides the input image into a regular grid and makes each cell predict two bounding boxes. Further iterations of the method, YOLOv2~\cite{Redmon_2017_CVPR_YOLO2} and YOLOv3\cite{redmon2018yolov3} built upon it by using anchor boxes, a better backbone network and several other tweaks. SSD~\cite{liu2016ssd} addresses the multi-scale detection problem by combining feature maps at multiple spatial resolutions and then applying anchor boxes to look for objects. RetinaNet~\cite{lin2018focal} uses an FPN (Feature pyramid network)~\cite{Lin_2017_CVPR_FPN} to produce a multi-scale pyramid of features and applies a set of anchor boxes followed by non-maximal suppression to find objects. CornerNet~\cite{law2018cornernet} uses the Hourglass network~\cite{newell2016stacked} paired with corner pooling layers to detect a set of top-left corners and bottom-right corners, and combines them with a learned embedding. Keypoint Triplets~\cite{duan2019centernet} builds upon CornerNet by improving the corner pooling layers, and by also detecting a center keypoint to validate each object. Objects as Points~\cite{zhou2019objectsaspoints} detects an object as a center keypoint and regresses the size of the object to find the bounding box.

\textbf{Attention mechanisms in object detection} have been around for a while. In Geometric Proposal for Faster R-CNN~\cite{amin2017geometric}, the authors re-rank the proposals of the region proposal network depending on a geometric estimation of the scene, outperforming the standard Faster R-CNN by a large margin. Their geometric estimation of the scene is mostly based on vehicle scale. The HAT~\cite{wu2019hierarchical} method uses a hierarchical attention mechanism that first trains a part-specific attention model. Then an LSTM models the relations between those parts, making it a part-aware detector. FG-BR Net~\cite{fu2019foreground} uses background subtraction methods to produce a foreground image that is fed as another input to the network. They also introduce a feedback process from the detection outputs to the background subtraction to keep static objects in the foreground image. Compared to these models, our attention mechanism is simple, elegant and fast. Furthermore, we do not need background subtraction at inference, only during the training phase.  

\section{Proposed Method}

\begin{figure*}[t]
\begin{center}
\includegraphics[width=\linewidth]{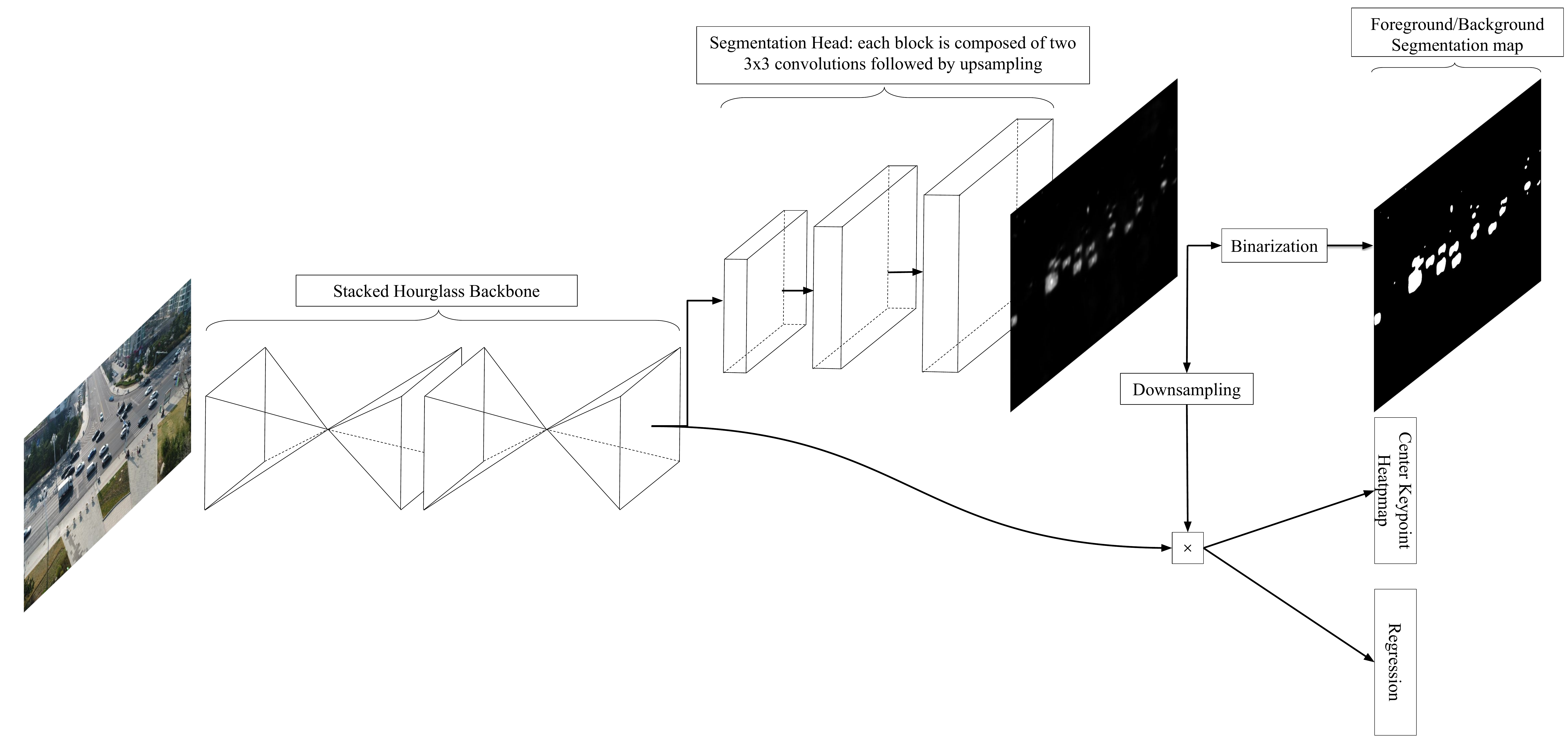}
\end{center}
   \caption{Overview of SpotNet: the input image first passes through a double-stacked hourglass network; the segmentation head then produces an attention map that multiplies the final feature map of the backbone network; the final center keypoint heatmap is then produced as well as the size and coordinate offset regressions for each object. }
\label{cn-seg}
\end{figure*}

Figure~\ref{cn-seg} shows a detailed overview of our complete model. 

\subsection{Base network}

Our method is based upon CenterNet~\cite{zhou2019objectsaspoints}, not to be confused with the homonym method CenterNet, or keypoint triplets~\cite{duan2019centernet}. This method trains a backbone network to recognize the center point of objects by assigning the center pixel of a box to be the ground-truth center and gives a reduced loss for other close points. The width and height of the bounding box are regressed, as well as the coordinate offset of the box (to compensate from the error caused by the smaller spatial resolution of the output). The final output is thus a center point heatmap for each possible label, an object size for each point, and an offset for each point, the size and offset being label agnostic. 

\subsection{Multi-task Learning}

Our main idea is to train a network to perform multiple tasks, to make it better for at least one of the tasks. In our case, we train a network to perform segmentation of objects of interest while performing bounding box detection, thus making the shared parameters more generic and less prone to overfitting. To do this, we add a two-class (foreground/background) segmentation head to the network and train this head with semi-supervised annotations from training datasets (more details in subsection~\ref{subsec:annot}).  

The added segmentation head takes as input a feature map that has been reduced by a factor of four in terms of spatial dimension when compared to the input. It consists of three 3 $\times$ 3 convolutions, with upsampling layers in between. The channel dimension is reduced to 1 in the last convolution, thus resulting in a segmentation map that is the same width and height as the input, with a single channel. The loss, $L_{seg}$, used to train this head is the binary cross-entropy, given by

\begin{equation}
    L_{seg} = -\frac{1}{N}\sum_{i=1}^{N} y_i*log(x_i) + (1-y_i)*log(1-x_i),
    \label{bce}
\end{equation}

where $y_i$ is the annotation label for sample $i$, $x_i$  its predicted label by the network and $N$ the number of samples. We found out during our experiments that it works better than the initial mean squared error loss that we had tried initially.  

\subsection{Self-Attention Mechanism}

To further benefit from our learned segmentation map, we implement a simple yet effective self-attention mechanism within the network. Once we obtain our segmentation map, we downsample it by a factor of 4 to reduce it to the spatial dimension of the original feature map. To attenuate the response at locations unlikely to contain an object of interest, we multiply every channel of the feature map with our segmentation map, thus reducing the probability of false positives in irrelevant areas. 

\subsection{Semi-Supervised annotations}
\label{subsec:annot}

\begin{figure}[ht]
\begin{center}
\includegraphics[width=1\linewidth]{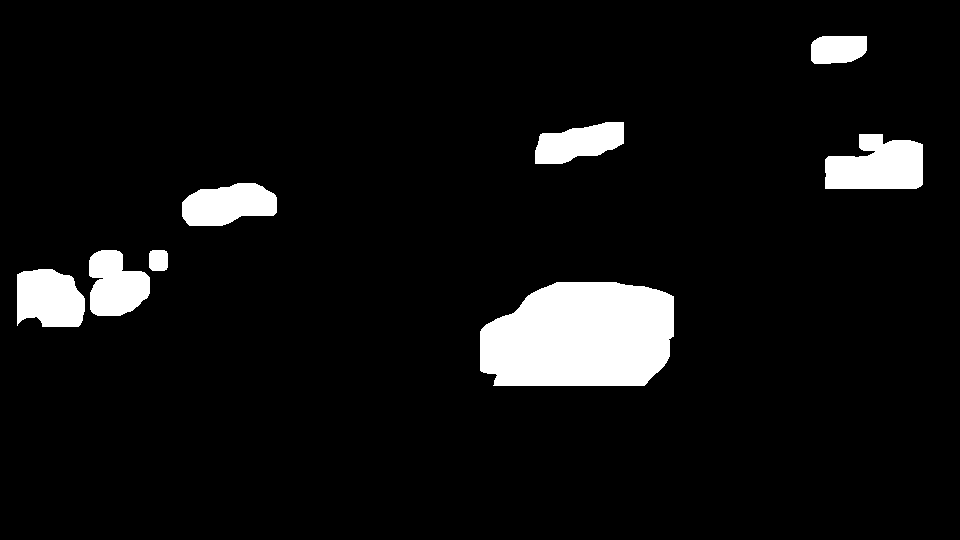}
\end{center}
   \caption{Example of semi-supervised annotations on UA-DETRAC~\cite{Wen2015Tracking} produced by PAWCS~\cite{st2015self_pawcs} and the intersection with the ground-truth bounding boxes.}
\label{annot}
\end{figure}

To train our model to produce foreground/background segmentation maps, we had to produce semi-supervised pixel-wise segmentation annotations. To do that, we took advantage of having access to full video sequences, despite training and evaluating on a single frame at a time. For the fixed camera video sequences, we used the background subtraction method PAWCS~\cite{st2015self_pawcs}. Since background subtraction is not designed to work with a moving background, for the moving camera video sequences, we used Farneback optical flow~\cite{farneback2003two} followed by some basic image processing and a threshold on motion magnitude. For both automatic two-class segmentation results, we then do an intersection with the ground-truth bounding boxes for each frame to reduce noise and to obtain pixel-wise segmentation annotations only for the object categories to detect. All other object categories, not inside ground-truth training bounding boxes, are therefore labelled as background. This results in fairly good foreground/background segmentation maps, with sometimes squared corners at one or more sides, due to the intersection with bounding boxes, as can be seen in Figure~\ref{annot}. In our experiments, we find that not only are these non-perfect segmentation annotations good enough to train good attention maps, it also allows our segmentation head to produce segmentation maps comparable to good unsupervised foreground/background segmentation methods. 

It should be noted that although our method requires videos for training to obtain the semi-supervised segmentation annotations, once trained, it can be applied to single images. 

\subsection{Training for multiple tasks}

To adapt the training loss of the whole network, we added the binary cross-entropy loss of our segmentation head (equation \ref{bce}) to the original CenterNet loss. The center point heatmap loss $L_{heat}$ is calculated with the focal loss~\cite{lin2018focal}, and the losses for the regressions for the offset $L_{off}$ and width/heigth $L_{WH}$ are formulated as L1 losses as in the original paper~\cite{zhou2019objectsaspoints}. The total loss $L_{tot}$ is given by

\begin{equation}
    L_{tot} = L_{heat} + L_{off} + L_{seg} + 0.1 * L_{WH}. 
    \label{loss}
\end{equation}
The total loss is thus the sum of all losses, with the width and height regression having less weight than the others, 0.1 compared to 1.

\section{Experiments}

\subsection{Datasets}

\begin{figure}[ht]
\begin{center}
\includegraphics[width=1\linewidth]{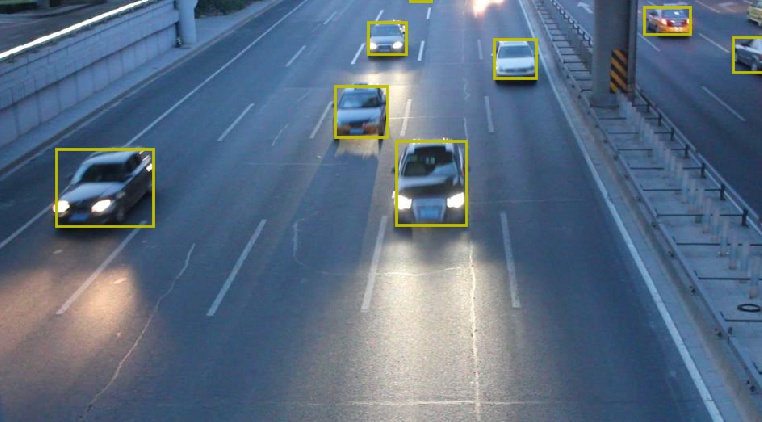}
\end{center}
   \caption{Sample from UA-DETRAC with the ground-truth bounding boxes in yellow.}
\label{ua-detrac}
\end{figure}

\begin{figure}[ht]
\begin{center}
\includegraphics[width=1\linewidth]{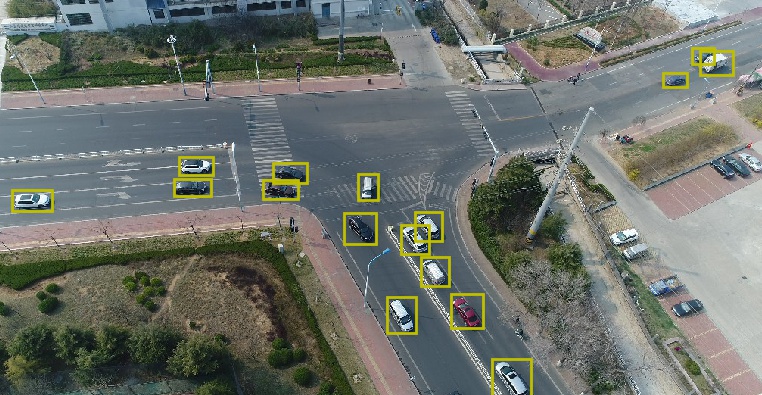}
\end{center}
   \caption{Sample from UAVDT with the ground-truth bounding boxes in yellow.}
\label{UAVDT}
\end{figure}

To validate the effectiveness of our method, we trained and evaluated it against other state-of-the-art methods on two datasets of traffic scenes, namely UA-DETRAC\cite{Wen2015Tracking} and UAVDT\cite{du2018unmanned}. Figure~\ref{ua-detrac} and Figure~\ref{UAVDT} show example frames of UA-DETRAC and UAVDT respectively, with their ground-truth. These two datasets were captured with very different settings, UA-DETRAC being filmed with a fixed camera for every scene, and UAVDT with a moving camera. Both datasets have pre-determined test sets, and we used a subset of the training data to do the validation.

Evaluation is done using the Matlab code provided by the authors of both datasets. A strict training and validation protocol was followed and the testing data was never seen by the network before the final evaluation. The performance measure used for evaluation is the mAP, the mean Average Precision, with a minimum IOU of 0.7 between inferred and ground-truth bounding boxes. The minimum IOU is the minimum overlap of a bounding box with the ground-truth to be considered a true detection. The IOU is computed as the intersection of the boxes divided by the union of the boxes. The mean average precision is the mean of the average precisions for all classes for multiple values of recall, ranging from 0 to 1 with small steps, typically of 0.1.  

\subsection{Implementation Details}

We used the stacked hourglass network as our backbone because it shows the best performance for keypoint estimation. This network is composed of modules of downsampling and convolutions followed by upsampling and convolutions with skip connections in an encoder-decoder fashion. For our experiments, we use the Hourglass-104 version as in~\cite{law2018cornernet} which stacks two encoder-decoder modules. We implemented the model in PyTorch 0.4.1 using Cuda 10.0. Experiments were run on a workstation with 32 GB of RAM and a NVIDIA GTX 1080Ti GPU. The Github repository for this project is \url{https://github.com/hu64/SpotNet}.

\subsection{Object detection results}

The experimental results are shown in Table~\ref{resultsuadetrac} for UA-DETRAC and in Table~\ref{results-UAVDT} for UAVDT. We outperform our baseline, CenterNet, by a very significant margin on both datasets while being the state-of-art results on both datasets as well. The results are very coherent, showing approximately the same percentage of improvement over CenterNet on both datasets, the absolute value on UAVDT being smaller.  

For UA-DETRAC, not only do we outperform all previously published results, we do so in every category, showing the benefit of our self-attention mechanism based on multi-task learning. Moreover, the improvements are particularly impressive for the category hard and cloudy, meaning that our model is particularly good for hard examples. It is interesting to note that the improvement for the easy category is very small, due to the mAP values being already very high. Nonetheless, improvement is consistent across all categories. At the moment of writing, our model outperforms every published result on this dataset, including ensemble models from challenges~\cite{lyu2018ua}.

The UAVDT dataset is more difficult than UA-DETRAC due to its high density of small vehicles and aerial point of view, but the percentage of improvement remains consistent. Our model also outperforms every published result on this dataset by a very significant margin. 

\begin{table*}[ht]
\footnotesize
\setlength\tabcolsep{3pt}
\def\arraystretch{1.5}
\centering
\caption{Results on the UA-DETRAC dataset~\cite{Wen2015Tracking}. 3D-DETNet results are from~\cite{3D_detnet_li20183d}, and others results are reported as in the results section of the UA-DETRAC website (\textbf{Boldface}: best result, \textit{Italic}: indicates our baseline).}
\label{resultsuadetrac}
\begin{tabular}{c|c|c|c|c|c|c|c|c}
Model & Overall & Easy & Medium & Hard & Cloudy & Night & Rainy & Sunny \\
\hline
\hline
SpotNet (ours) & \textbf{86.80}\% & \textbf{97.58}\% & \textbf{92.57}\% & \textbf{76.58}\% & \textbf{89.38}\% & \textbf{89.53}\% & \textbf{80.93}\% & \textbf{91.42}\% \\
\hline 
\textit{CenterNet}\cite{duan2019centernet} & 83.48\% & 96.50\% & 90.15\% & 71.46\% & 85.01\% & 88.82\% & 77.78\% & 88.73\% \\
\hline
FG-BR\_Net~\cite{fu2019foreground} & 79.96\% & 93.49\% & 83.60\% & 70.78\% & 87.36\% & 78.42\% & 70.50\% & 89.8\%\\
\hline 
HAT~\cite{wu2019hierarchical} & 78.64\% & 93.44\% & 83.09\% & 68.04\% & 86.27\% & 78.00\% & 67.97\% & 88.78\% \\
\hline
GP-FRCNNm~\cite{amin2017geometric} & 77.96\% & 92.74\% & 82.39\% & 67.22\% & 83.23\% & 77.75\% & 70.17\% & 86.56\% \\
\hline 
R-FCN~\cite{RFCN_NIPS2016_6465} & 69.87\% &    93.32\% &    75.67\% &    54.31\% &    74.38\% &    75.09\% &    56.21\% &    84.08\% \\
\hline
EB~\cite{EB_wang2017evolving} & 67.96\%    & 89.65\% &    73.12\% & 53.64\% & 72.42\% & 73.93\% & 53.40\% & 83.73\% \\
\hline 
Faster R-CNN~\cite{ren2015faster} & 58.45\% &    82.75\% &    63.05\% &    44.25\% &    66.29\% &    69.85\% &    45.16\% &    62.34\% \\
\hline 
YOLOv2~\cite{Redmon_2017_CVPR_YOLO2} & 57.72\% &    83.28\% &    62.25\% &    42.44\% &    57.97\% &    64.53\% &    47.84\% &    69.75\% \\
\hline 
RN-D~\cite{perreault2019road} & 54.69\% &    80.98\% &    59.13\% &    39.23\% &    59.88\% &    54.62\% &    41.11\% &    77.53\% \\
\hline 
3D-DETnet~\cite{3D_detnet_li20183d} & 53.30\% &    66.66\% &    59.26\% &    43.22\% &    63.30\% &    52.90\% &    44.27\% &    71.26\% \\
\end{tabular}
\vspace{-4mm}
\end{table*}

\begin{table}[ht]
\footnotesize
\setlength\tabcolsep{3pt}
\def\arraystretch{1.5}
\centering
\caption{Results on the UAVDT~\cite{du2018unmanned} dataset (\textbf{Boldface}: best result, \textit{Italic}: indicates our baseline).}
\label{results-UAVDT}
\begin{tabular}{c|c}
Model & Overall \\
\hline
\hline
SpotNet (Ours)& \textbf{52.80}\%\\
\hline
\textit{CenterNet}\cite{duan2019centernet}& 51.18\%\\
\hline
Wang \etal ~\cite{wang2019learning} & 37.81\%\\
\hline
R-FCN~\cite{RFCN_NIPS2016_6465} &34.35\%\\
\hline
SSD~\cite{liu2016ssd} & 33.62\%\\
\hline
Faster-RCNN~\cite{ren2015faster} & 22.32\%\\
\hline
RON~\cite{kong2017ron} & 21.59\%\\
\end{tabular}
\vspace{-4mm}
\end{table}

\subsection{Foreground/Background segmentation results}

\begin{figure*}[ht]%
    \centering
    \subfloat[Input Image]{\includegraphics[width=0.15\linewidth]{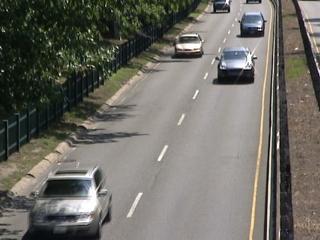}\label{highway_in}}
    \subfloat[Ground-truth]{\includegraphics[width=0.15\linewidth]{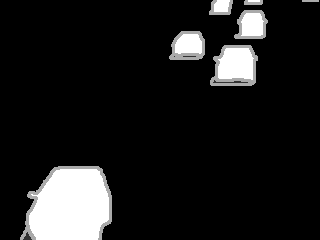}\label{highway_gt}}
    \subfloat[SpotNet (ours)]{\includegraphics[width=0.15\linewidth]{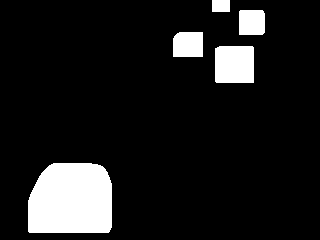}\label{highway_spotnet}}
    \subfloat[PAWCS~\cite{st2015self_pawcs}]{\includegraphics[width=0.15\linewidth]{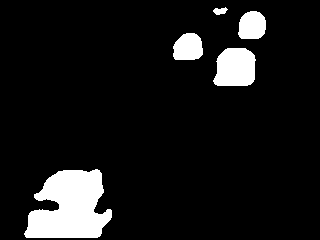}\label{highway_pawcs}}
    \subfloat[SGMM~\cite{evangelio2012splitting_sgmm}]{\includegraphics[width=0.15\linewidth]{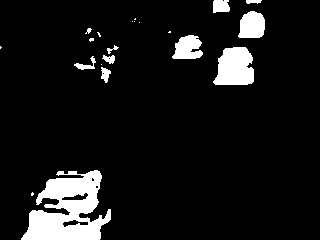}\label{highway_sgmm}}
    \subfloat[GMM~\cite{stauffer1999adaptive_gmm}]{\includegraphics[width=0.15\linewidth]{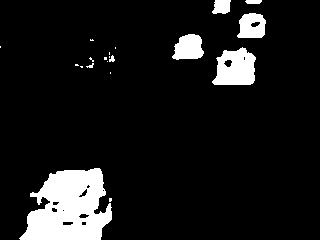}\label{highway_gmm}}
    \vspace{-2.2em}
    \subfloat[]{\includegraphics[width=0.15\linewidth]{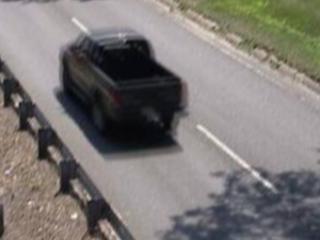}\label{traffic_in}}
    \subfloat[]{\includegraphics[width=0.15\linewidth]{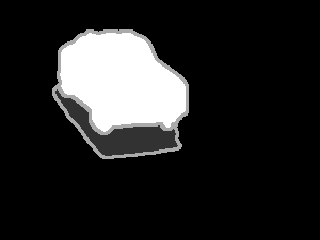}\label{traffic_gt}}
    \subfloat[]{\includegraphics[width=0.15\linewidth]{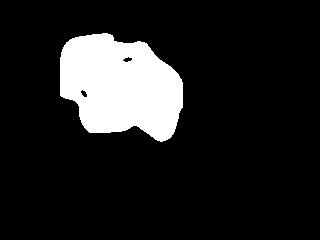}\label{traffic_spotnet}}
    \subfloat[]{\includegraphics[width=0.15\linewidth]{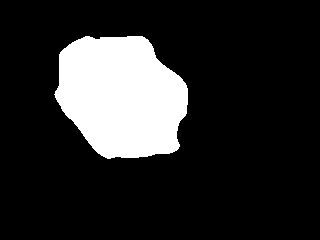}\label{traffic_pawcs}}
    \subfloat[]{\includegraphics[width=0.15\linewidth]{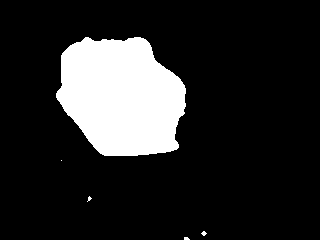}\label{traffic_sgmm}}
    \subfloat[]{\includegraphics[width=0.15\linewidth]{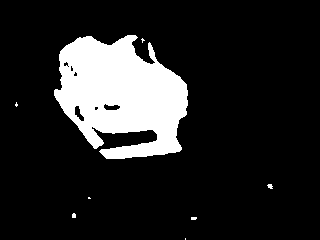}\label{traffic_gmm}}
    \vspace{-2.2em}
    \subfloat[]{\includegraphics[width=0.15\linewidth]{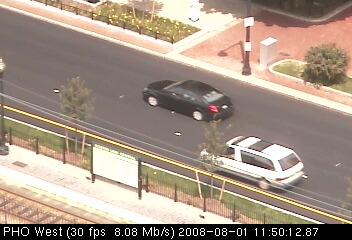}\label{boulevard_in}}
    \subfloat[]{\includegraphics[width=0.15\linewidth]{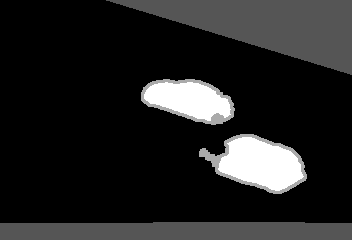}\label{boulevard_gt}}
    \subfloat[]{\includegraphics[width=0.15\linewidth]{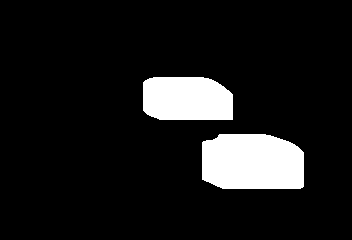}\label{boulevard_spotnet}}
    \subfloat[]{\includegraphics[width=0.15\linewidth]{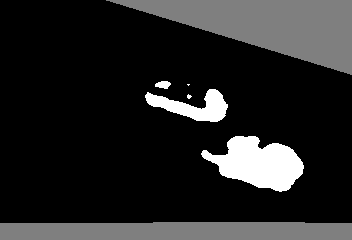}\label{boulevard_pawcs}}
    \subfloat[]{\includegraphics[width=0.15\linewidth]{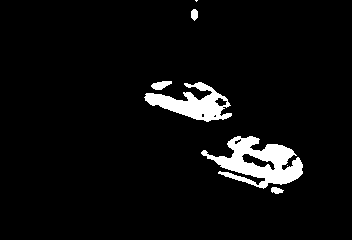}\label{boulevard_sgmm}}
    \subfloat[]{\includegraphics[width=0.15\linewidth]{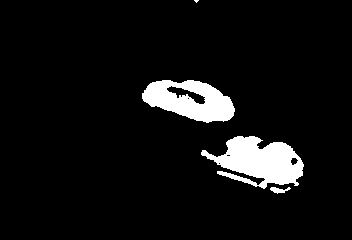}\label{boulevard_gmm}}
    \caption{Example of foreground/background segmentation maps obtained with several segmentation methods. First row: frame 1015 of ``highway'', second row: frame 967 of ``traffic'', third row: frame 883 of ``boulevard''.}
    \label{seg-example}
\end{figure*}

Although that was not our principal objective, it is nonetheless interesting to see how we do on specialized foreground/background benchmarks. To produce results, we used our best model trained on UA-DETRAC and ran it on three sequences of the changedetection.net dataset~\cite{goyette2012changedetection} containing only vehicles (because UA-DETRAC includes only annotations for vehicles). To obtain the foreground, we took the attention maps produced by our network, applied a binary threshold and then masked the resulting image with the bounding boxes detected by our network to remove noise. We can see in table~\ref{results-changedetection} that our method produces competitive results, although we do not quite reach state-of-the-art foreground/background performance. Figure~\ref{seg-example} shows qualitative results on a few frames. Our method does not always fit the object boundaries very well. This is expected since the training annotations are imperfect. Nevertheless, we outperform several classical methods, at no additional cost when producing bounding boxes. It is important to note that a limitation of our model is that it must be trained on the objects we want to segment.  

\begin{table}[ht]
\footnotesize
\setlength\tabcolsep{3pt}
\def\arraystretch{1.5}
\centering
\caption{Results on the changedetection.net~\cite{goyette2012changedetection} dataset. Results are averaged for sequences ``highway'', ``traffic'' and ``boulevard'' (\textbf{Boldface}: best result).}
\label{results-changedetection}
\begin{tabular}{c|c}
Model & Average F-Measure \\
\hline
\hline
PAWCS~\cite{st2015self_pawcs} & \textbf{0.872} \\
\hline
SuBSENSE~\cite{st2014subsense}  & 0.831 \\
\hline
SpotNet (Ours) & 0.806 \\
\hline
SGMM~\cite{evangelio2012splitting_sgmm}  & 0.766 \\
\hline
KNN~\cite{zivkovic2006efficient_knn}  & 0.731 \\
\hline
GMM~\cite{stauffer1999adaptive_gmm}  & 0.709 \\
\end{tabular}
\vspace{-4mm}
\end{table}

\captionsetup[subfigure]{labelformat=empty, position=top}

\section{Discussion}

\subsection{Ablation study}

\begin{figure}[ht]
\begin{center}
\includegraphics[width=1\linewidth]{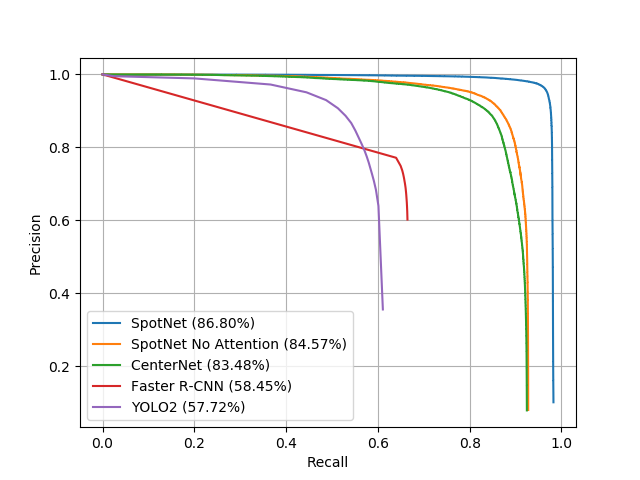}
\end{center}
   \caption{Precision/Recall curve of our model compared with a variant and other methods.}
\label{prec-rec}
\end{figure}

To detail the contribution of each part of our model, we conducted an ablation study on UA-DETRAC. Table~\ref{ablation} shows that even though multi-task learning helps, the biggest contribution comes from combining our attention process with it. To further understand the contribution of each part, we draw the precision/recall curve (Figure~\ref{prec-rec}) compared to several other methods on UA-DETRAC. On this curve, we can note that the multi-task learning by itself (SpotNet No Attention) helps to be more precise, but does not help to detect more objects, i.e.\ to reach improved values of recall. On the other hand, the attention mechanism does both, it helps to be even more precise for the same values of recall (fewer false positives), and it also allows the model to detect more and reach significantly higher values of recall. 

\begin{table*}[ht]
\footnotesize
\setlength\tabcolsep{3pt}
\def\arraystretch{1.5}
\centering
\caption{Ablation study on the UA-DETRAC~\cite{Wen2015Tracking} dataset.}
\label{ablation}
\begin{tabular}{c|c|c|c|c|c|c|c|c|c}
Attention & Multi-Task  & Overall & Easy & Medium & Hard & Cloudy & Night & Rainy & Sunny \\
\hline
\hline
\checkmark & \checkmark & \textbf{86.80}\% & \textbf{97.58}\% & \textbf{92.57}\% & \textbf{76.58}\% & \textbf{89.38}\% & \textbf{89.53}\% & \textbf{80.93}\% & \textbf{91.42}\% \\
\hline 
 & \checkmark & 84.57\% & 96.72\% & 90.85\% & 73.16\% & 86.53\% & 88.76\% & 78.84\% & 90.10\% \\
\hline 
 & & 83.48\% & 96.50\% & 90.15\% & 71.46\% & 85.01\% & 88.82\% & 77.78\% & 88.73\% \\
\end{tabular}
\vspace{-4mm}
\end{table*}

Since the network is looking for keypoints on the whole image, it is natural that concentrating the search on learned foreground pixels will increase the probability that the keypoints found belong to the objects of interest, thus reducing the rate of false positives. Furthermore, the experiments show that this increases recall because the network can concentrate on useful information.  

It is expected that learning the segmentation task jointly with the object detection task can be mutually beneficial since both tasks have a large overlap in what needs to be learned. The main difference is that object detection needs to separate instances, while segmentation needs a more precise border around the objects. We show that semi-supervised annotations are good enough for our purpose, and multi-task learning by itself, based on those annotations, improves precision. 

\subsection{Limitations of our Model}

 One of the limitations of our model is the fact that it needs semi-supervised annotations to be trained properly. However, we believe that in most real-world applications, video sequences are available and we can thus run background subtraction or optical flow to generate them. In other cases, pre-trained semantic segmentation methods could be used to obtain the desired annotations. 

\section{Conclusion}

In this paper, we presented a novel multi-task model equipped with a self-attention process, and we trained it with semi-supervised annotations. We show that these improvements allow us to reach state-of-the-art performance on two traffic scenes datasets with different settings. We argue that not only does this improve accuracy by a large margin, it also provides instance segmentations of the road users almost at no cost. 

\section*{Acknowledgment}
We acknowledge the support of the Natural Sciences and Engineering Research Council of Canada (NSERC), [RDCPJ 508883 - 17], and the support of Genetec.

\bibliographystyle{IEEEtran}
\bibliography{bib.bib}

\begin{thebibliography}{10}
\providecommand{\url}[1]{#1}
\csname url@samestyle\endcsname
\providecommand{\newblock}{\relax}
\providecommand{\bibinfo}[2]{#2}
\providecommand{\BIBentrySTDinterwordspacing}{\spaceskip=0pt\relax}
\providecommand{\BIBentryALTinterwordstretchfactor}{4}
\providecommand{\BIBentryALTinterwordspacing}{\spaceskip=\fontdimen2\font plus
\BIBentryALTinterwordstretchfactor\fontdimen3\font minus
  \fontdimen4\font\relax}
\providecommand{\BIBforeignlanguage}[2]{{%
\expandafter\ifx\csname l@#1\endcsname\relax
\typeout{** WARNING: IEEEtran.bst: No hyphenation pattern has been}%
\typeout{** loaded for the language `#1'. Using the pattern for}%
\typeout{** the default language instead.}%
\else
\language=\csname l@#1\endcsname
\fi
#2}}
\providecommand{\BIBdecl}{\relax}
\BIBdecl

\bibitem{law2018cornernet}
H.~Law and J.~Deng, ``Cornernet: Detecting objects as paired keypoints,'' in
  \emph{Proceedings of the European Conference on Computer Vision (ECCV)},
  2018, pp. 734--750.

\bibitem{duan2019centernet}
K.~Duan, S.~Bai, L.~Xie, H.~Qi, Q.~Huang, and Q.~Tian, ``Centernet: Keypoint
  triplets for object detection,'' in \emph{Proceedings of the IEEE
  International Conference on Computer Vision}, 2019, pp. 6569--6578.

\bibitem{zhou2019objectsaspoints}
X.~Zhou, D.~Wang, and P.~Kr{\"a}henb{\"u}hl, ``Objects as points,'' \emph{arXiv
  preprint arXiv:1904.07850}, 2019.

\bibitem{xu2015show}
K.~Xu, J.~Ba, R.~Kiros, K.~Cho, A.~Courville, R.~Salakhudinov, R.~Zemel, and
  Y.~Bengio, ``Show, attend and tell: Neural image caption generation with
  visual attention,'' in \emph{International conference on machine learning},
  2015, pp. 2048--2057.

\bibitem{vaswani2017attention}
A.~Vaswani, N.~Shazeer, N.~Parmar, J.~Uszkoreit, L.~Jones, A.~N. Gomez,
  {\L}.~Kaiser, and I.~Polosukhin, ``Attention is all you need,'' in
  \emph{Advances in neural information processing systems}, 2017, pp.
  5998--6008.

\bibitem{du2018unmanned}
D.~Du, Y.~Qi, H.~Yu, Y.~Yang, K.~Duan, G.~Li, W.~Zhang, Q.~Huang, and Q.~Tian,
  ``The unmanned aerial vehicle benchmark: Object detection and tracking,'' in
  \emph{Proceedings of the European Conference on Computer Vision (ECCV)},
  2018, pp. 370--386.

\bibitem{rcnn_Girshick_2014_CVPR}
R.~Girshick, J.~Donahue, T.~Darrell, and J.~Malik, ``Rich feature hierarchies
  for accurate object detection and semantic segmentation,'' in \emph{The IEEE
  Conference on Computer Vision and Pattern Recognition (CVPR)}, 2014.

\bibitem{uijlings2013selective}
J.~R. Uijlings, K.~E. Van De~Sande, T.~Gevers, and A.~W. Smeulders, ``Selective
  search for object recognition,'' \emph{International journal of computer
  vision}, vol. 104, no.~2, pp. 154--171, 2013.

\bibitem{Girshick_2015_ICCV_fast}
R.~Girshick, ``Fast r-cnn,'' in \emph{The IEEE International Conference on
  Computer Vision (ICCV)}, 2015.

\bibitem{ren2015faster}
S.~Ren, K.~He, R.~Girshick, and J.~Sun, ``Faster r-cnn: Towards real-time
  object detection with region proposal networks,'' in \emph{Advances in neural
  information processing systems}, 2015, pp. 91--99.

\bibitem{RFCN_NIPS2016_6465}
J.~Dai, Y.~Li, K.~He, and J.~Sun, ``R-fcn: Object detection via region-based
  fully convolutional networks,'' in \emph{Advances in Neural Information
  Processing Systems 29}.\hskip 1em plus 0.5em minus 0.4em\relax Curran
  Associates, Inc., 2016, pp. 379--387.

\bibitem{cai2018cascade}
Z.~Cai and N.~Vasconcelos, ``Cascade r-cnn: Delving into high quality object
  detection,'' in \emph{Proceedings of the IEEE conference on computer vision
  and pattern recognition}, 2018, pp. 6154--6162.

\bibitem{redmon2016you_yolo}
J.~Redmon, S.~Divvala, R.~Girshick, and A.~Farhadi, ``You only look once:
  Unified, real-time object detection,'' in \emph{Proceedings of the IEEE
  conference on computer vision and pattern recognition}, 2016, pp. 779--788.

\bibitem{Redmon_2017_CVPR_YOLO2}
J.~Redmon and A.~Farhadi, ``Yolo9000: Better, faster, stronger,'' in \emph{The
  IEEE Conference on Computer Vision and Pattern Recognition (CVPR)}, 2017.

\bibitem{redmon2018yolov3}
------, ``Yolov3: An incremental improvement,'' \emph{arXiv preprint
  arXiv:1804.02767}, 2018.

\bibitem{liu2016ssd}
W.~Liu, D.~Anguelov, D.~Erhan, C.~Szegedy, S.~Reed, C.-Y. Fu, and A.~C. Berg,
  ``Ssd: Single shot multibox detector,'' in \emph{European conference on
  computer vision}.\hskip 1em plus 0.5em minus 0.4em\relax Springer, 2016, pp.
  21--37.

\bibitem{lin2018focal}
T.-Y. Lin, P.~Goyal, R.~Girshick, K.~He, and P.~Doll{\'a}r, ``Focal loss for
  dense object detection,'' \emph{IEEE transactions on pattern analysis and
  machine intelligence}, 2018.

\bibitem{Lin_2017_CVPR_FPN}
T.-Y. Lin, P.~Dollar, R.~Girshick, K.~He, B.~Hariharan, and S.~Belongie,
  ``Feature pyramid networks for object detection,'' in \emph{The IEEE
  Conference on Computer Vision and Pattern Recognition (CVPR)}, 2017.

\bibitem{newell2016stacked}
A.~Newell, K.~Yang, and J.~Deng, ``Stacked hourglass networks for human pose
  estimation,'' in \emph{European conference on computer vision}.\hskip 1em
  plus 0.5em minus 0.4em\relax Springer, 2016, pp. 483--499.

\bibitem{amin2017geometric}
S.~Amin and F.~Galasso, ``Geometric proposals for faster r-cnn,'' in \emph{2017
  14th IEEE International Conference on Advanced Video and Signal Based
  Surveillance (AVSS)}.\hskip 1em plus 0.5em minus 0.4em\relax IEEE, 2017, pp.
  1--6.

\bibitem{wu2019hierarchical}
S.~Wu, M.~Kan, S.~Shan, and X.~Chen, ``Hierarchical attention for part-aware
  face detection,'' \emph{International Journal of Computer Vision}, vol. 127,
  no. 6-7, pp. 560--578, 2019.

\bibitem{fu2019foreground}
Z.~Fu, Y.~Chen, H.~Yong, R.~Jiang, L.~Zhang, and X.-S. Hua, ``Foreground gating
  and background refining network for surveillance object detection,''
  \emph{IEEE Transactions on Image Processing}, vol.~28, no.~12, pp.
  6077--6090, 2019.

\bibitem{st2015self_pawcs}
P.-L. St-Charles, G.-A. Bilodeau, and R.~Bergevin, ``A self-adjusting approach
  to change detection based on background word consensus,'' in \emph{2015 IEEE
  winter conference on applications of computer vision}.\hskip 1em plus 0.5em
  minus 0.4em\relax IEEE, 2015, pp. 990--997.

\bibitem{farneback2003two}
G.~Farneb{\"a}ck, ``Two-frame motion estimation based on polynomial
  expansion,'' in \emph{Scandinavian conference on Image analysis}.\hskip 1em
  plus 0.5em minus 0.4em\relax Springer, 2003, pp. 363--370.

\bibitem{Wen2015Tracking}
L.~Wen, D.~Du, Z.~Cai, Z.~Lei, M.-C. Chang, H.~Qi, J.~Lim, M.-H. Yang, and
  S.~Lyu, ``{ UA-DETRAC: A New Benchmark and Protocol for Multi-Object
  Detection and Tracking},'' \emph{arXiv CoRR}, vol. abs/1511.04136, 2015.

\bibitem{lyu2018ua}
S.~Lyu, M.-C. Chang, D.~Du, W.~Li, Y.~Wei, M.~Del~Coco, P.~Carcagn{\`\i},
  A.~Schumann, B.~Munjal, D.-H. Choi \emph{et~al.}, ``Ua-detrac 2018: Report of
  avss2018 \& iwt4s challenge on advanced traffic monitoring,'' in \emph{2018
  15th IEEE International Conference on Advanced Video and Signal Based
  Surveillance (AVSS)}.\hskip 1em plus 0.5em minus 0.4em\relax IEEE, 2018, pp.
  1--6.

\bibitem{3D_detnet_li20183d}
S.~Li and F.~Chen, ``3d-detnet: a single stage video-based vehicle detector,''
  in \emph{Third International Workshop on Pattern Recognition}, vol.
  10828.\hskip 1em plus 0.5em minus 0.4em\relax International Society for
  Optics and Photonics, 2018, p. 108280A.

\bibitem{EB_wang2017evolving}
L.~Wang, Y.~Lu, H.~Wang, Y.~Zheng, H.~Ye, and X.~Xue, ``Evolving boxes for fast
  vehicle detection,'' in \emph{2017 IEEE International Conference on
  Multimedia and Expo (ICME)}.\hskip 1em plus 0.5em minus 0.4em\relax IEEE,
  2017, pp. 1135--1140.

\bibitem{perreault2019road}
H.~Perreault, G.-A. Bilodeau, N.~Saunier, and P.~Gravel, ``Road user detection
  in videos,'' \emph{arXiv preprint arXiv:1903.12049}, 2019.

\bibitem{wang2019learning}
T.~Wang, R.~M. Anwer, H.~Cholakkal, F.~S. Khan, Y.~Pang, and L.~Shao,
  ``Learning rich features at high-speed for single-shot object detection,'' in
  \emph{Proceedings of the IEEE International Conference on Computer Vision},
  2019, pp. 1971--1980.

\bibitem{kong2017ron}
T.~Kong, F.~Sun, A.~Yao, H.~Liu, M.~Lu, and Y.~Chen, ``Ron: Reverse connection
  with objectness prior networks for object detection,'' in \emph{IEEE
  Conference on Computer Vision and Pattern Recognition}, vol.~1, 2017, p.~2.

\bibitem{goyette2012changedetection}
N.~Goyette, P.-M. Jodoin, F.~Porikli, J.~Konrad, and P.~Ishwar,
  ``Changedetection. net: A new change detection benchmark dataset,'' in
  \emph{2012 IEEE computer society conference on computer vision and pattern
  recognition workshops}.\hskip 1em plus 0.5em minus 0.4em\relax IEEE, 2012,
  pp. 1--8.

\bibitem{st2014subsense}
P.-L. St-Charles, G.-A. Bilodeau, and R.~Bergevin, ``Subsense: A universal
  change detection method with local adaptive sensitivity,'' \emph{IEEE
  Transactions on Image Processing}, vol.~24, no.~1, pp. 359--373, 2014.

\bibitem{evangelio2012splitting_sgmm}
R.~H. Evangelio, M.~P{\"a}tzold, and T.~Sikora, ``Splitting gaussians in
  mixture models,'' in \emph{2012 IEEE Ninth international conference on
  advanced video and signal-based surveillance}.\hskip 1em plus 0.5em minus
  0.4em\relax IEEE, 2012, pp. 300--305.

\bibitem{zivkovic2006efficient_knn}
Z.~Zivkovic and F.~Van Der~Heijden, ``Efficient adaptive density estimation per
  image pixel for the task of background subtraction,'' \emph{Pattern
  recognition letters}, vol.~27, no.~7, pp. 773--780, 2006.

\bibitem{stauffer1999adaptive_gmm}
C.~Stauffer and W.~E.~L. Grimson, ``Adaptive background mixture models for
  real-time tracking,'' in \emph{Proceedings. 1999 IEEE Computer Society
  Conference on Computer Vision and Pattern Recognition (Cat. No PR00149)},
  vol.~2.\hskip 1em plus 0.5em minus 0.4em\relax IEEE, 1999, pp. 246--252.

\end{thebibliography}

\end{document}